# Institutional-Level Monitoring of Immune Checkpoint Inhibitor IrAEs Using a Novel Natural Language Processing Algorithmic Pipeline

Michael Shapiro M.D.[1,2], Herut Dor Ph.D.[2], Anna Gurevich-Shapiro M.D.[2,3,4], Tal Etan M.D.[2,5], Ido Wolf M.D.[2,5]

[1]Department of Internal Medicine T, Tel Aviv Sourasky Medical Center.
[2]Sackler Faculty of Medicine, Tel Aviv University, Tel Aviv, Israel.
[3]Department of Systems Immunology, Weizmann Institute of Science, Rehovot, Israel.
[4]Division of Hematology, Tel Aviv Sourasky Medical Center, Tel Aviv, Israel.
[5]Oncology Division, Tel Aviv Sourasky Medical Center, Tel Aviv, Israel.

Word count: 3996

Corresponding Author:

Michael Shapiro, MD MSc, ORCID: 0000-0001-5943-6974

Department of Internal Medicine T

Tel Aviv Sourasky Medical Center

Weizmann St 6, Tel Aviv

Tel: +972-545278620

E-mail: mikehpg@gmail.com

## Abstract (Words – 241)

Background: Immune checkpoint inhibitors (ICIs) have revolutionized cancer treatment but can result in severe immune-related adverse events (IrAEs). Monitoring IrAEs on a large scale is essential for personalized risk profiling and assisting in treatment decisions.

Methods: In this study, we conducted an analysis of clinical notes from patients who received ICIs at the Tel Aviv Sourasky Medical Center. By employing a Natural Language Processing algorithmic pipeline, we systematically identified seven common or severe IrAEs. We examined the utilization of corticosteroids, treatment discontinuation rates following IrAEs, and constructed survival curves to visualize the occurrence of adverse events during treatment.

Results: Our analysis encompassed 108,280 clinical notes associated with 1,635 patients who had undergone ICI therapy. The detected incidence of IrAEs was consistent with previous reports, exhibiting substantial variation across different ICIs. Treatment with corticosteroids varied depending on the specific IrAE, ranging from 17.3% for thyroiditis to 57.4% for myocarditis. Our algorithm demonstrated high accuracy in identifying IrAEs, as indicated by an area under the curve (AUC) of 0.89 for each suspected note and F1 scores of 0.87 or higher for five out of the seven IrAEs examined at the patient level.

Conclusions: This study presents a novel, large-scale monitoring approach utilizing deep neural networks for IrAEs. Our method provides accurate results, enhancing understanding of detrimental consequences experienced by ICI-treated patients. Moreover, it holds potential for monitoring other medications, enabling comprehensive post-marketing surveillance to identify susceptible populations and establish personalized drug safety profiles.

# OBJECTIVE

Immune checkpoint inhibitors (ICI) have revolutionized treatment in some of the most difficult to treat malignancies. However, their ability to activate the immune system is a double-edged sword that can target healthy tissues, causing a wide array of immune-related adverse events (IrAEs). These adverse events lead to significant morbidity and mortality associated with ICI and thus are an important consideration in treatment decisions.

In the clinical trial setting, adverse events associated with treatment are closely monitored by dedicated study teams using specialized questionnaires.[1,2] However, this method is labor-intensive and not routinely implemented in the busy everyday oncology clinics. Additionally, immune-mediated adverse events are often diagnosed by specialists such as endocrinologists, pulmonologists, gastroenterologists, and cardiologists, rather than oncologists. As ICIs are a relatively new class of medications, and guidelines for the identification and management of IrAEs are still being developed,[3] active monitoring of IrAEs at the institutional level is essential for ensuring patient safety and improving the standard of care.

An efficient and accurate system for monitoring ICIs adverse events on a large scale would provide extensive and consistent data, allowing for a more comprehensive understanding of drug safety that is not subject to human error or under-reporting. This would promote equal and inclusive surveillance, providing data relevant to the local patients' population and populations under-represented in clinical trials, creating personalized safety profiles, and informing clinical decision-making. Furthermore, such a system would be invaluable for regulatory agencies to detect harmful drugs faster and for pharmaceutical companies to improve drug development.

Electronic medical records (EMRs), a staple of the digital transformation of medicine, provide an opportunity to solve this problem. EMRs contain structured information including International Classification of Disease (ICD) diagnosis and lab results that is easy to access but requires the health provider to enter it intentionally and consistently into the system, a laborious task that is often skipped at the busy daily clinic. Thus, missing structured data is common. The unstructured data is mostly comprised of the free text entered by the physicians as part of their routine work and mandatory documentation, inseparable of modern medicine, thus containing much of the information known about the patient. The unstructured nature of the data has traditionally posed a challenge for automated analysis. However, recent advancements in Natural Language Processing (NLP) have enabled the development of language models that can effectively analyze text. NLP techniques, such as language translation, text summarization, and question answering, are used in a variety of applications, such as chatGPT,[4] and have become increasingly prevalent in other domains. However, applying NLP in the medical field has presented unique challenges due to the specialized terminology, need of domain specific models and privacy and security challenges.[5]

Herein, we present an innovative approach for institutional-level monitoring of ICIs-mediated IrAEs, using an NLP algorithm. This is a three-step process: comprised of identifying potential adverse events mentions, validation of those references using a combination of two deep learning approaches and finally clustering to ensure the specificity of findings. Our success in applying the algorithm hospital-wide and achieving results similar to those reported in clinical trials indicate a potential for implementing similar methods for post-marketing and real-world data collection regarding adverse events of other classes of medications.

# MATERIALS AND METHODS

**Data source and study population**

The population of this study consisted of all patients who received treatment with one of six immune checkpoint inhibitors: atezolizumab, avelumab, durvalumab, pembrolizumab, nivolumab or the ipilimumab-nivolumab combination at Sourasky Medical Center between January 2015 and August 2021. The data used in this study, including prescription information and clinical notes, were extracted from the hospitals EMRs and included any follow up, admission letters, or discharge notes written between the first administration of the immunotherapy and up to three months after the last administration. The text was primarily written in Hebrew, with a small number of English words included.

To ensure an accurate evaluation of the algorithm, 20% of the patients enrolled in the study were randomly selected for use as test data. This selection process was done at the patient level to prevent label leakage, as similar wording can appear in the records of the same patient due to doctors' tendency to repeat phrases from previous visits. This would have resulted in a bias if the notes regarding the same patient were used for both training the algorithm and testing it.

**Adverse event definition and data annotation**

This study focused on the following IrAEs: pneumonitis, hepatitis, thyroiditis, colitis, myocarditis, dermatitis, and myasthenia gravis. These events were selected by the oncological department for their importance as common or severe adverse events prominent in daily practice. This also allowed for the evaluation of the system's ability to detect both prevalent and rare adverse events.

The National Cancer Institute's Common Terminology Criteria for Adverse Events version 5.0 was used as a guide for our definition of adverse events. IrAEs Diagnoses were determined based solely on the treating physician's written notes in the free text in the EMRs, as the structured ICD codes were infrequently used for adverse event documentation. The annotation of the data was conducted by a single physician (M.S.) with the assistance of the oncology team to resolve any uncertainties. This approach ensured consistency and accuracy in the identification of adverse events.

**Survival analysis**

To investigate the timing of immune-related adverse events (IrAEs) across different drug regimens, we plotted survival curves using data from both the training and test cohorts. The event date for each IrAE was defined as the first mention of the adverse event. Kaplan-Meier curves were used to depict the IrAEs free survival for the four commonly used drugs: Nivolumab, Pembrolizumab, Atezolizumab, and the combination of Ipilimumab and Nivolumab. To improve the readability of the visual representation, confidence intervals have been omitted due to significant overlap.

**Corticosteroid treatment**

To verify the accuracy of our algorithm and evaluate the incidence of severe (grade 2-4) IrAEs, we analyzed the EMRs of the study population for corticosteroid treatment regimens that were administered or prescribed within two weeks after the algorithm first detected an IrAE, including prednisone or methylprednisolone with a dose of 1 to 2 mg/kg a day.[6] This analysis encompassed both inpatient care as well as treatment prescribed in outpatient clinics.

**ICI treatment discontinuation**

A retrospective analysis of patient data was performed to evaluate the discontinuation of ICI treatment following the development of an IrAE. Patients were considered to have discontinued ICI treatment if they received their last dose of ICI within one month prior to the onset of the IrAE and did not receive any subsequent treatment.

**The algorithm workflow**

The algorithm pipeline in our study uses a three-step process to detect and confirm IrAEs, it is presented in a step-by-step fashion in supplementary figure S1.

1. <u>Identification of potential adverse events:</u> All notes were scanned for words in Hebrew or English that resemble the words used for the IrAEs of interest. We used the Levenshtein distance (also known as Edit distance) as a similarity measure as implemented by the "thefuzz" python package (https://github.com/seatgeek/thefuzz).[7] The acceptable difference was set high to achieve near 100% sensitivity at this stage while still filtering out much of the data in a computationally efficient manner. For example, in the search for "pneumonitis" words that were flagged in English included: 'pneumonirtis', 'pnumonitis', 'pnemonitis', 'pneumnuitis', 'pneumonytis' and 'pneumocistis'.
2. <u>Validation of the IrAEs references:</u> All notes flagged by the first step were then classified according to whether they contained information indicating that the patient was experiencing a specific adverse event and it was not attributed to causes other than ICIs, i.e. radiation. To achieve this, we extracted five words before and after the word identified in the first step, and used these 11 words to construct a sentence for classification.

We examined two methods as a classification model for this task:

*Method 1:* We used pre-trained transformer based models, trained on Hebrew text. These included XLM-Roberta,[8] BERT,[9] and AlephBERT.[10] The first two models were trained as a multi-language model with Hebrew only as part of the training, AlephBERT is a pre-trained language model specifically for Hebrew, thus potentially able to perform better for out data. As the medical language is different from the training data used to pre-train these models, we performed "domain adaptation" using unsupervised learning employing random EMR notes to achieve better results. The models were then trained in a process of transfer learning for the classification of IrAEs. The "Transformers" python package was used for this method.[11]

*Method 2:* We constructed a new language model using the FastText word embedding by unsupervised learning on random notes from the EMRs.[12] The embedding created by this method uses sub-words rather than entire word, thus it is more robust for spelling mistakes common in medical text. An LSTM based neural network, useful for small datasets, was then used for classification.[13] The model was then trained using supervised learning on the training data.

An ensemble of the two models was then constructed by averaging on the probability output of the two models. Due to utilizing two very distinct methods we hypnotized this will boost performance.

3. Clustering: To mitigate false positives, we employed a data aggregation approach. Initially, all references pertaining to the same IrAE within a clinical note were collectively examined, considering the note positive for that IrAE if at least one reference was positive. Subsequently, a patient was classified as experiencing a particular IrAE if they had at least two positive notes for that IrAE at any point during their treatment, and the date of the IrAE occurrence was assigned as the date of the first positive note for that IrAE.

**Evaluation**

We evaluated each step of the algorithm separately:

1. To evaluate the sensitivity of detecting IrAEs using word similarity, we randomly selected 1000 notes from our database. Given the limited ability to scan all available notes, we specifically used notes written in the oncology clinics as this increased the chances of capturing IrAEs in the randomly selected batch. Each note was examined both by the algorithm for any words similar to the IrAEs in question and manually scanned by the authors (M.S) for any mentions of the IrAEs. It should be noted that except for the notes in this evaluation, notes that were not flagged by the algorithm as containing potential IrAE were not examined by hand. This step primarily aimed to minimize the data required for evaluation in the subsequent stage, with our focus on maintaining high sensitivity while decreasing the number of sentences retained for analysis. In addition, our ability to conduct a reliable probability analysis was constrained by the limited number of notes we were able to manual scan. Consequently, a more thorough evaluation of event detection was deferred to the second stage of the pipeline.
2. The patients in our cohort were randomly split, with 80% of patients serving as the training cohort and 20% serving as the test cohort. All suspected IrAE occurrences flagged by the previous step of the pipeline in clinical notes regarding the training cohort patients were part of the training data and similarly for the test cohort. Each clinical note could contain references of several IrAEs and each reference was considered independently. The training set consisted of all IrAEs suspected references with the neighboring words extracted to form the input for the algorithm following the methodology outlined in the previous section. The same process was applied to form the test set. Each reference was individually labeled by a physician to determine whether it contained a positive reference to an IrAE. We used Receiver Operating Characteristic Area Under the Curve (ROC AUC) as the measure for classifying a single sentence as containing a reference stating that the patient has the adverse event and that no other etiology for the IrAE other than ICI treatment was mentioned. This was named the "sentence level performance". The ensemble model was evaluated the same way.

3. The results of the best model in terms of AUC (whether a single or ensemble model) were clustered as mentioned previously. We compared the results of this step for each IrAE monitored to aggregation using the manual labeling in terms of sensitivity, specificity, F1 score and accuracy. These results were referred to as the "patient-level performance".

**Scalability**

In order to evaluate the scalability of our algorithm, we conducted a follow-up study by monitoring the patients in our cohort from the end of the follow-up in the previous step up to October 2022. The start of the additional follow up for each patient was three months after the last dose of ICI or August 2021, the earlies. The end of the follow-up period for each patient was determined by the date of the last note recorded in the EMR. We analyzed all notes created during this time period, using the previously described algorithm to identify instances of IrAEs. These results were compared to previously detected IrAEs for each patient. Newly identified IrAEs were investigated for high-dose corticosteroid treatment. We evaluated the number of detected IrAEs, the rate of corticosteroid usage compared to the physician-audited data from the previous sections.

Additionally, we examined the runtime of the algorithm. The hardware used for this evaluation was an i9-10885H CPU, 32GB RAM, and an Nvidia Quadro RTX 4000 8GB RAM GPU.

**Code availability**

The code used in this study is available from: https://github.com/DrMikeSh/Immuneverse_public

# RESULTS

**Medical Notes and Patient's Characteristics**

The study included 1,635 patients treated with the predetermined ICIs by the oncology clinics of Tel Aviv Sourasky Medical Center between 2015 and 2021, and analyzed a total of 108,280 unique medical notes. The sample was divided into a training and validation group of 1308 patients, and a test group of 327 patients. Table 1 presents demographic information for the study participants. The majority of patients in the cohort received immunotherapy as monotherapy, and approximately 50% of them received ICIs as a first line treatment. The patients were diagnosed with a diverse range of malignancies, with lung and urological cancers being the most common.

**Table 1:** Cohort characteristics

|  | **Train** | **Test** | **All patients** |
|---|---|---|---|
| N | 1308 | 327 | 1635 |
| Combination treatment | 22.7% | 23.5% | 22.8% |
| **Line of therapy** | | | |
| First | 50.4% | 48.0% | 49.9% |
| Second | 27.6% | 31.8% | 28.5% |
| Third | 13.9% | 11.6% | 13.4% |
| More advanced | 8.1% | 8.5% | 8.2% |
| Age (Mean +/- SD) | 65.9 ± 13.1 | 65.6 ± 13.1 | 65.7 ± 13.1 |
| Female (%) | 39.4% | 41.5% | 41.1% |
| **Treated malignancies** | | | |
| Lung | 368 (27.9%) | 109 (34.28%) | 477 (29.14%) |
| Urological malignancies | 172 (13.04%) | 57 (17.92%) | 229 (13.99%) |
| Skin | 180 (13.65%) | 26 (8.18%) | 206 (12.58%) |
| Brain and nervous system | 115 (8.72%) | 30 (9.43%) | 145 (8.86%) |
| Gynecological malignancies | 61 (4.62%) | 18 (5.66%) | 79 (4.83%) |
| Oral, larynx and pharynx malignancies | 70 (5.31%) | 13 (4.09%) | 83 (5.07%) |
| Stomach | 62 (4.7%) | 16 (5.03%) | 78 (4.76%) |
| Breast | 44 (3.34%) | 10 (3.14%) | 54 (3.3%) |
| Colon | 61 (4.62%) | 11 (3.46%) | 72 (4.4%) |
| Liver and bile tract | 38 (2.88%) | 7 (2.20%) | 45 (2.75%) |
| Other GI malignancies | 32 (2.43%) | 4 (1.26%) | 36 (2.2%) |
| Other | 105 (8.11%) | 26 (8.18%) | 131 (8.12%) |

**Immune related Adverse Events**

Table 2 presents the prevalence of IrAEs for each of the immune checkpoint inhibitors evaluated in the study. Pneumonitis and thyroiditis were the most frequently observed adverse events, with prevalence rates reaching up to 20%. The Ipilimumab and Nivolumab combination resulted in the highest rate of IrAEs, while Nivolumab monotherapy was found to be the safest treatment among those examined.

**Table 2:** Prevalence of IrAEs in the entire cohort for each checkpoint inhibitor

| Drug | N | Pneumonitis | Hepatitis | Thyroiditis | MG | Dermatitis | Colitis | Myocarditis |
|---|---|---|---|---|---|---|---|---|
| Atezolizumab | 122 | 13 (10.7%) | 5 (4.1%) | 6 (4.9%) | 0 (0.0%) | 3 (2.5%) | 1 (0.8%) | 4 (3.3%) |
| Avelumab | 11 | 1 (9.1%) | 0 (0.0%) | 1 (9.1%) | 0 (0.0%) | 0 (0.0%) | 0 (0.0%) | 0 (0.0%) |
| Durvalumab | 58 | 12 (20.7%) | 1 (1.7%) | 3 (5.2%) | 1 (1.7%) | 1 (1.7%) | 0 (0.0%) | 3 (5.2%) |
| Pembrolizumab | 926 | 88 (9.5%) | 21 (2.3%) | 43 (4.6%) | 5 (0.5%) | 23 (2.5%) | 30 (3.2%) | 22 (2.4%) |
| Nivolumab | 424 | 37 (8.7%) | 7 (1.7%) | 11 (2.6%) | 2 (0.5%) | 5 (1.2%) | 9 (2.1%) | 5 (1.2%) |
| Ipilimumab + Nivolumab | 94 | 14 (14.9%) | 12 (12.8%) | 17 (18.1%) | 2 (2.1%) | 4 (4.3%) | 6 (6.4%) | 3 (3.2%) |

The IrAE-free survival in our study population is depicted using Kaplan-Meier curves in Figure 1 for the four most commonly used immunotherapy regimens. The combination of Ipilimumab and Nivolumab was associated with earlier onset of adverse events, with most occurring within the first year (n = 45, 95.7%) of treatment. For Pembrolizumab, the most commonly used immunotherapy agent, the data was sufficient to demonstrate a semi-linear rate of incidence over time for most adverse events, with a unique rate for each IrAE. Hepatitis, in contrast to other IrAEs, occurred exclusive during the first year of treatment for all ICIs examined (n = 42, 97.7%), with one case occurring after 373 days.

**Corticosteroids use**

The frequency of high-dose corticosteroid administration in the two weeks following initial identification of an IrAE is presented in Table 3. The rate varied between different ICIs and across IrAE types. Overall, myasthenia gravis (75%), hepatitis (58.7%), and myocarditis (56.8%) were most commonly treated with high-dose corticosteroids, while thyroiditis was primarily managed without steroid treatment (18.3%). Patients who developed IrAEs while receiving the combination regimen of Ipilimumab + Nivolumab were the most likely to received corticosteroid treatment. The overall rate of steroid treatment across all regimens and IrAEs was 32.8%.

**Table 3:** Corticosteroid treatment following the diagnosis of IrAEs in the entire cohort

| Drug | N | Pneumonitis | Hepatitis | Thyroiditis | MG | Dermatitis | Colitis | Myocarditis |
|---|---|---|---|---|---|---|---|---|
| Atezolizumab | 122 | 9 (69.2%) | 5 (100%) | 0 | 0 | 0 | 0 | 2 (50%) |
| Avelumab | 11 | 0 | 0 | 0 | 0 | 0 | 0 | 0 |
| Durvalumab | 58 | 3 (25%) | 1 (100%) | 0 | 1 (100%) | 0 | 0 | 2 (66.7%) |
| Pembrolizumab | 926 | 36 (40.1%) | 13 (61.9%) | 4 (9.3%) | 4 (80%) | 6 (26.1%) | 14 (46.7%) | 11 (50%) |
| Nivolumab | 424 | 12 (32.4%) | 1 (14.3%) | 4 (36.4%) | 0 | 1 (20%) | 2 (22.2%) | 4 (80%) |
| Ipilimumab + Nivolumab | 94 | 9 (64.3%) | 7 (58.3%) | 5 (29.4%) | 1 (50%) | 2 (50%) | 2 (33.3%) | 2 (66.6%) |
| Mean steroid treatment rate (%) | | 42.1% | 58.7% | 18.3% | 75.0% | 28.1% | 40.0% | 56.8% |

The percentage in parentheses represents the proportion of patients who received corticosteroids among those with identified IrAEs.

**ICI treatment discontinuation**

The frequency of treatment discontinuation potentially associated with IrAE is presented in Table S2. The results showed that the highest frequency of treatment discontinuation was observed in patients with myasthenia gravis (40%), followed by those with myocarditis (35.1%) and hepatitis (30.4%). Interestingly, patients receiving the combination treatment of Ipilimumab + Nivolumab had a lower frequency of treatment discontinuation compared to those receiving other ICIs, despite receiving corticosteroids at higher rates. This potentially suggests a better response to the corticosteroids. The overall treatment discontinuation rate was 21.1%.

**Algorithm performance**

IrAEs detection

We identified 43 notes that contained mention of IrAEs through manual review, and our algorithm successfully detected all of them. The low threshold used in this initial stage of the algorithm allowed us to achieve near 100% sensitivity, while later steps of the algorithm were implemented to enhance specificity.

Sentence level performance

The AlephBERT and fastText-LSTM models demonstrated the highest ROC-AUC scores at around 0.84. To further improve performance, we created an ensemble model that combined the output of the two best models for each sample, resulting in an ROC-AUC score of 0.889 (Figure 2). The ensemble model was used for the rest of the study.

Patient level performance

The final performance metrics for detecting patients with each IrAE, including sensitivity, specificity, F1 score, and accuracy, are presented in Table 4. The algorithm performed well for most IrAEs, achieving an F1 score of more than 87% for five of the seven IrAEs examined. However, there were higher rates of false positives for colitis. The results of the ensemble algorithm's predictions were compared to the manually reviewed results in Table S1.

**Table 4:** The performance of the algorithm on the test cohort for each of the IrAEs examined.

| IrAE | Sensitivity | Specificity | F1 score | Accuracy |
|---|---|---|---|---|
| Colitis | 88.89% | 97.48% | 64.00% | 99.69% |
| Dermatitis | 75.00% | 99.69% | 80.00% | 99.39% |
| Hepatitis | 100.00% | 99.37% | 88.89% | 99.39% |
| Myasthenia Gravis | 100.00% | 100.00% | 100.00% | 100.00% |
| Myocarditis | 100.00% | 100.00% | 100.00% | 100.00% |
| Pneumonitis | 90.91% | 97.96% | 86.96% | 99.08% |
| Thyroiditis | 91.67% | 100.00% | 95.65% | 99.69% |

Scalability

The additional average follow-up period for patients was 409 days during which 175,350 notes were recorded. Our algorithm identified 98 new cases of IrAE among 83 patients. The most frequently observed adverse events were pneumonitis and thyroiditis, in line with our earlier findings. Detailed results for each drug are presented in Table S3. The rate of steroid treatment for IrAEs, at 32.6%, was similar to the rate observed in the manually reviewed data (32.8%). The total computational runtime for the algorithm was 190.75 seconds.

## DISCUSSION

In this study, we present a novel tool for identifying and monitoring IrAEs on a large scale using deep learning-based algorithm for analyzing free-text notes in EMRs. Our results demonstrate that the algorithmic pipeline achieves IrAE reporting rates similar to those of clinical trials, with a high proportion of associated corticosteroid treatment and ICI treatment discontinuation, supporting its validity. To the best of our knowledge, this is the first tool to achieve this goal on such a large scale, using real world data, providing a valuable tool for monitoring and management of IrAEs in clinical practice.

ICIs have revolutionized cancer treatment by enhancing anti-tumor immunity, however, they can also cause excessive activation of the immune system. The immune adverse events of checkpoint inhibitors have been primarily studied through data from clinical trials and meta-analysis while real-world data remains scarce.[14] Most real-world studies use the FDA Adverse Event Reporting System (FAERS) system or VigiBase (World Health Organization's global database of individual case safety reports) as source of data.[15] These registries are based on physician reporting, thus focus on severe AEs and mortality, and likely underestimate the overall prevalence.[16] A large-scale analysis using insurance claims data revealed an increase in immune adverse events compared to matched patients receiving chemotherapy but used ICD codes, many of them are not specific.[17] To date, the real extent of immune adverse events is not known.[18]

Previous studies that utilized natural language processing in the medical domain have focused on extracting symptoms from free text, including heart failure diagnostic criteria,[19] detecting multiple sclerosis cases,[20] and detecting clusters of COVID-19[21]. Several studies examined the detection of adverse events mainly from a methodological viewpoint and in small cohorts.[22–25] In the field of oncology, several studies were able to detect symptoms from clinical notes with significant accuracy in small datasets.[26,27] The detection of IrAEs in clinical notes has remained a challenge,[28] with one study proposing a method to detect notes containing any IrAE with poorer results in identifying the specific adverse event.[29] To address this challenge, a robust algorithm is needed that can effectively analyze the diverse and voluminous clinical data that characterizes real-world scenarios. This includes the ability to handle numerous clinical notes from both hospitalization and clinic visits, as well as the capability to scale for institutional-wide evaluations.

In this study, we propose a scalable algorithmic pipeline for evaluating both common and rare IrAEs among all oncological patients treated with ICIs in the Sourasky medical center over a five-year period. To achieve this, we conducted a comprehensive analysis of clinical notes from all departments and clinics within the hospital. Additionally, to ensure the reliability of the data, the algorithm was designed to exclude IrAEs that are attributed to causes other than ICI treatment by the treating physician. We used a three step deep neural network algorithm that demonstrated high sensitivity and a low rate of false positive results achieving an F1 score >0.87 for most of the adverse events examined. This performance was achieved with minimal computational runtime and despite being applied to Hebrew language data, which lacks adequate NLP resources and pre-trained models for transfer learning.[10] We speculate that training a similar model in a language with better resources, such as English, may lead to even better performance.

The results of our algorithm demonstrated a higher rate of pneumonitis, with 9-15% of patients receiving immunotherapy suffering from the adverse event, compared to the 1-5% reported in previous studies.[30] Myocarditis was also more common in our study (2.3%) compared to previous reports.[31] We found thyroiditis rates of 3-18% of patients, similar to previous studies,[32,33] as well as lower incidence of colitis,[34] hepatitis,[35] and dermatitis.[36] We believe the discrepancies could be explained by our use of more stringent criteria for diagnosis of IrAEs, where our classification relied on physician confirmation of the condition as an IrAE, rather than relying solely on symptoms such as rash and diarrhea while on ICI treatment; Moreover, some IrAEs were under-reported in previous years, most notably myocarditis,[37] and our algorithm provides a solution for this surveillance challenge.

Prior investigations have established an early onset pattern for IrAEs, with the median occurrence within 16 weeks of treatment,[38] reaching its peak in the initial 4 weeks.[39] Other studies have identified instances where the onset extends beyond one year of treatment.[40] Our current study indicates that, with the exception of hepatitis, which predominantly manifested within the first year, the remaining IrAEs examined showed a nearly constant rate of occurrence, with survival curves adopting a linear trajectory over a 3-year follow-up period. This observed trend was particularly prominent among patients administered Pembrolizumab, the most frequently utilized ICI in our investigation. It should be noted that, while thyroiditis among Nivolumab-treated patients was predominantly observed within the first year, those receiving Pembrolizumab were diagnosed with thyroiditis even after more than two years of treatment. We attribute the disparity between our results and prior reports to the scale of our cohort as well as an investigation of several ICIs, as earlier studies often presented outcomes based on smaller patient populations, shorter follow-up durations (e.g. the duration of a clinical trial) and for a single ICI. This underscores that there is much to be investigated regarding IrAEs temporal dynamics, especially as ICI treatment becomes increasingly widespread and patients undergo prolonged treatment regimens.

The rate of high-dose corticosteroid administration observed in our cohort in relation to IrAE identification was consistent with previous reports,[41] likely reflecting the presence of grade 2 or 4 IrAEs.[42] The rate of ICI discontinuation, at 21%, was higher than what has been reported in clinical trials,[43] highlighting the significance of real-world data in evaluating the impact of IrAEs on patients' treatment outcomes and prognosis. Moreover, Ipilimumab and Nivolumab combination treatment had a low rate of treatment discontinuation despite having the highest rate of IrAEs and corticosteroid treatment. These results suggest that different ICIs have varying toxicity profiles and further research is needed to determine the best option for patients in terms of toxicity and risk of drug discontinuation, not just the ICI anti-cancer potency.

The methodology presented in our study, easily adjustable for other medications and adverse events, provides the possibility to achieve safety monitoring on a large scale, avoiding the reliance on subjective reporting.[44] This system offers up-to-date, accurate and consistent results that are less subject to human error and are not resource intensive. Therefore, this system can boost pharmacovigilance by delivering and inclusive safety data for various age and ethnic groups in the population that are underrepresented in current clinical trials. Thus, providing a more complete and nuanced picture of drug safety and a personalized risk-benefit consideration to inform patient selection. Additionally, this method can be invaluable to regulatory agencies to detect harmful drugs early, especially important in accelerated or conditional approval of new drugs.[45] Finally, this system can provide information for drug companies for developing safer treatments, i.e. as part of the next generation of ICIs.

This study has several limitations. First, the study relies on the treating physician to make the diagnosis of the adverse event and attribute it to the ICI treatment. However, as most adverse events presentation can be caused by other conditions, the physician's evaluation here is essential. Second, the study was conducted in a single center and the test data used, though as separate as possible from the training data through patient based splitting, was still from the same institution. Third, we did not evaluate the model for detection of adverse events it was not trained for, but, as we demonstrated efficiency both for rare conditions as Myasthenia gravis and common as pneumonitis, we believe that if needed the model can be successfully adjusted to detect any other adverse event. Finally, in the analysis of corticosteroid treatment and ICI discontinuation, 6% and 10% of patients, respectively, had more than one IrAE detected for the first time during the relevant period and it was not possible to determine which IrAE was more clinically significant. Although both events were recorded, we believe their limited number had a minimal impact on the overall conclusions drawn from the data.

# CONCLUSION

In conclusion, our multi-step algorithmic pipeline has proven to be an efficient and accurate solution for detecting IrAEs among oncology patients treated with ICIs at the Tel Aviv Sourasky Medical Center. This study provides valuable insights into the incidence, temporal dynamics, and severity of IrAEs, including the need for corticosteroids and discontinuation of treatment, based on real-world data on a scale and in detail that has not been achieved before. Our findings suggest that similar models could be used for large-scale, real-time post-marketing surveillance, providing a patient-tailored safety profile, early warning of severe adverse events, ultimately leading to improved prognosis and quality of life for these patients.

# DECLERATIONS

Ethics approval: This study was approved by the Tel Aviv Sourasky medical center Ethics Review Board which waived written informed consent (Reference ID 1031-20- TLV).

Data availability: The data used in this study is comprised of clinical notes from patients treated with immune checkpoint inhibitors at the Tel Aviv Sourasky Medical Center and is considered confidential and protected by patient privacy regulations. The data cannot be made publicly available due to ethical and legal constraints.

Code availability: The code used for this study is available at
https://github.com/DrMikeSh/Immuneverse_public

Competing Interests: Prof. Ido Wolf has provided lectures and consulting services to BMS, MSD, Roche, Astra, and Novartis, other authors declare that they have no competing interests.

Funding information: This study was funded by the Kahn Foundation ORION scholarship program.

Authors' Contributions: MS was responsible for the study design, data collection and analysis, development of the computational tools, and drafting of the manuscript. HD and AGS contributed to the study design and manuscript writing. TE provided valuable revision suggestions and guidance during the study. IW conceived the study idea, participated in its design, provided guidance and supervision, and co-wrote the manuscript.

Acknowledgements: The authors express their gratitude to the Health Quality and Patient Safety department of Tel Aviv Sourasky Medical Center for their invaluable support in this project.

# REFERENCES

1 Tang S-Q, Tang L-L, Mao Y-P, *et al.* The Pattern of Time to Onset and Resolution of Immune-Related Adverse Events Caused by Immune Checkpoint Inhibitors in Cancer: A Pooled Analysis of 23 Clinical Trials and 8,436 Patients. *Cancer Res Treat*. 2021;53:339–54.

2 Thapa B, Roopkumar J, Kim AS, *et al.* Incidence and clinical pattern of immune related adverse effects (irAE) due to immune checkpoint inhibitors (ICI). *https://doi.org/101200/JCO20193715_suppl.e14151*. 2019;37:e14151–e14151.

3 Allouchery M, Lombard T, Martin M, *et al.* Original research: Safety of immune checkpoint inhibitor rechallenge after discontinuation for grade ≥2 immune-related adverse events in patients with cancer. *Journal for Immunotherapy of Cancer*. 2020;8:1622.

4 Radford A, Wu J, Child R, *et al.* Language models are unsupervised multitask learners. *OpenAI blog*. 2019;1:9.

5 Adlung L, Cohen Y, Mor U, *et al.* Machine learning in clinical decision making. *Med*. 2021;2:642–65.

6 Brahmer JR, Lacchetti C, Schneider BJ, *et al.* Management of immune-related adverse events in patients treated with immune checkpoint inhibitor therapy: American society of clinical oncology clinical practice guideline. *Journal of Clinical Oncology*. 2018;36:1714–68.

7 Levenshtein VI, Levenshtein, I. V. Binary Codes Capable of Correcting Deletions, Insertions and Reversals. *SPhD*. 1966;10:707.

8 Conneau A, Khandelwal K, Goyal N, *et al.* Unsupervised Cross-lingual Representation Learning at Scale. 2019;132–5.

9 Devlin J, Chang MW, Lee K, *et al.* BERT: Pre-training of Deep Bidirectional Transformers for Language Understanding. *NAACL HLT 2019 - 2019 Conference of the North American Chapter of the Association for Computational Linguistics: Human Language Technologies - Proceedings of the Conference*. 2018;1:4171–86.

10 Seker A, Bandel E, Bareket D, *et al.* AlephBERT:A Hebrew Large Pre-Trained Language Model to Start-off your Hebrew NLP Application With. Published Online First: 8 April 2021. doi: 10.48550/arxiv.2104.04052

11 Wolf T, Debut L, Sanh V, *et al.* Transformers: State-of-the-Art Natural Language Processing. 2020;38–45.

12 Joulin A, Grave E, Bojanowski P, *et al.* Bag of Tricks for Efficient Text Classification. *15th Conference of the European Chapter of the Association for Computational Linguistics, EACL 2017 - Proceedings of Conference*. 2016;2:427–31.

13 Badjatiya P, Gupta S, Gupta M, *et al.* Deep learning for hate speech detection in tweets. *26th International World Wide Web Conference 2017, WWW 2017 Companion*. 2017;759–60.

14 Martins F, Sofiya L, Sykiotis GP, *et al.* Adverse effects of immune-checkpoint inhibitors: epidemiology, management and surveillance. *Nature Reviews Clinical Oncology 2019 16:9*. 2019;16:563–80.

15  Raschi E, Gatti M, Gelsomino F, *et al.* Lessons to be Learnt from Real-World Studies on Immune-Related Adverse Events with Checkpoint Inhibitors: A Clinical Perspective from Pharmacovigilance. *Targeted Oncology*. 2020;15:449–66.

16  Moslehi JJ, Salem JE, Sosman JA, *et al.* Increased reporting of fatal immune checkpoint inhibitor-associated myocarditis. *The Lancet*. 2018;391:933.

17  Wang F, Yang S, Palmer N, *et al.* Real-world data analyses unveiled the immune-related adverse effects of immune checkpoint inhibitors across cancer types. *npj Precision Oncology 2021 5:1*. 2021;5:1–11.

18  Postow MA, Sidlow R, Hellmann MD. Immune-Related Adverse Events Associated with Immune Checkpoint Blockade. *New England Journal of Medicine*. 2018;378:158–68.

19  Byrd RJ, Steinhubl SR, Sun J, *et al.* Automatic identification of heart failure diagnostic criteria, using text analysis of clinical notes from electronic health records. *International Journal of Medical Informatics*. 2014;83:983–92.

20  Chase HS, Mitrani LR, Lu GG, *et al.* Early recognition of multiple sclerosis using natural language processing of the electronic health record. *BMC medical informatics and decision making*. 2017;17:24.

21  Shapiro M, Landau R, Shay S, *et al.* Early detection of COVID-19 outbreaks using textual analysis of electronic medical records. *Journal of Clinical Virology*. 2022;155:105251.

22  Ju M, Nguyen NTH, Miwa M, *et al.* An ensemble of neural models for nested adverse drug events and medication extraction with subwords. *Journal of the American Medical Informatics Association*. 2020;27:22–30.

23  Dai H-J, Su C-H, Wu C-S. Adverse drug event and medication extraction in electronic health records via a cascading architecture with different sequence labeling models and word embeddings. *Journal of the American Medical Informatics Association*. 2020;27:47–55.

24  Boyce Jeremy; Miller, Taylor; Kane-Gill, Sandra L. RD; J. Automated Screening of Emergency Department Notes for Drug-Associated Bleeding Adverse Events Occurring in Older Adults. *Appl Clin Inform*. 2017;08:1022–30.

25  Li F, Liu W, Yu H. Extraction of Information Related to Adverse Drug Events from Electronic Health Record Notes: Design of an End-to-End Model Based on Deep Learning. *JMIR medical informatics*. 2018;6. doi: 10.2196/12159

26  Hong JC, Fairchild AT, Tanksley JP, *et al.* Natural language processing for abstraction of cancer treatment toxicities: accuracy versus human experts. *JAMIA Open*. 2021;3:513–7.

27  Lindvall C, Deng C-Y, Agaronnik ND, *et al.* Deep Learning for Cancer Symptoms Monitoring on the Basis of Electronic Health Record Unstructured Clinical Notes. *JCO Clinical Cancer Informatics*. Published Online First: 17 June 2022. doi: 10.1200/cci.21.00136

28  Reynolds KL, Arora S, Elayavilli RK, *et al.* Immune-related adverse events associated with immune checkpoint inhibitors: a call to action for collecting and sharing clinical trial and real-world data. *Journal for ImmunoTherapy of Cancer*. 2021;9:e002896.

29  Gupta S, Belouali A, Shah NJ, *et al.* Automated Identification of Patients With Immune-Related Adverse Events From Clinical Notes Using Word Embedding and Machine Learning. *JCO Clinical Cancer Informatics*. 2021;541–9.

30    Nishino M, Giobbie-Hurder A, Hatabu H, *et al.* Incidence of Programmed Cell Death 1 Inhibitor–Related Pneumonitis in Patients With Advanced Cancer: A Systematic Review and Meta-analysis. *JAMA Oncology*. 2016;2:1607–16.
31    Palaskas N, Lopez-Mattei J, Durand JB, *et al.* Immune Checkpoint Inhibitor Myocarditis: Pathophysiological Characteristics, Diagnosis, and Treatment. *Journal of the American Heart Association*. 2020;9. doi: 10.1161/JAHA.119.013757
32    Ryder M, Callahan M, Postow MA, *et al.* Endocrine-related adverse events following ipilimumab in patients with advanced melanoma: a comprehensive retrospective review from a single institution. *Endocrine-Related Cancer*. 2014;21:371–81.
33    Barroso-Sousa R, Barry WT, Garrido-Castro AC, *et al.* Incidence of Endocrine Dysfunction Following the Use of Different Immune Checkpoint Inhibitor Regimens: A Systematic Review and Meta-analysis. *JAMA Oncology*. 2018;4:173–82.
34    Champiat S, Lambotte O, Barreau E, *et al.* Management of immune checkpoint blockade dysimmune toxicities: a collaborative position paper. *Annals of Oncology*. 2016;27:559–74.
35    Sanjeevaiah A, Kerr T, Beg MS. Approach and management of checkpoint inhibitor-related immune hepatitis. *Journal of Gastrointestinal Oncology*. 2018;9:220.
36    Minkis K, Garden BC, Wu S, *et al.* The risk of rash associated with ipilimumab in patients with cancer: A systematic review of the literature and meta-analysis. *Journal of the American Academy of Dermatology*. 2013;69:e121–8.
37    Palaskas N, Lopez-Mattei J, Durand JB, *et al.* Immune Checkpoint Inhibitor Myocarditis: Pathophysiological Characteristics, Diagnosis, and Treatment. *Journal of the American Heart Association*. 2020;9. doi: 10.1161/JAHA.119.013757
38    Ramos-Casals M, Brahmer JR, Callahan MK, *et al.* Immune-related adverse events of checkpoint inhibitors. *Nat Rev Dis Primers*. 2020;6:38.
39    Kanjanapan Y, Day D, Butler MO, *et al.* Delayed immune-related adverse events in assessment for dose-limiting toxicity in early phase immunotherapy trials. *Eur J Cancer*. 2019;107:1–7.
40    Yoest JM. Clinical features, predictive correlates, and pathophysiology of immune-related adverse events in immune checkpoint inhibitor treatments in cancer: a short review. *Immunotargets Ther*. 2017;6:73.
41    Horvat TZ, Adel NG, Dang TO, *et al.* Immune-Related Adverse Events, Need for Systemic Immunosuppression, and Effects on Survival and Time to Treatment Failure in Patients With Melanoma Treated With Ipilimumab at Memorial Sloan Kettering Cancer Center. *Journal of Clinical Oncology*. 2015;33:3193.
42    Ramos-Casals M, Brahmer JR, Callahan MK, *et al.* Immune-related adverse events of checkpoint inhibitors. *Nature Reviews Disease Primers 2020 6:1*. 2020;6:1–21.
43    Khoja L, Day D, Wei-Wu Chen T, *et al.* Tumour- and class-specific patterns of immune-related adverse events of immune checkpoint inhibitors: a systematic review. *Annals of Oncology*. 2017;28:2377–85.
44    Palassin P, Faillie J-L, Coustal C, *et al.* Underreporting of Major Cardiac Adverse Events With Immune Checkpoint Inhibitors in Clinical Trials: Importance of Postmarketing Pharmacovigilance Surveys. *Journal of Clinical Oncology : Official Journal of the American Society of Clinical Oncology*. 2022;JCO2201603–JCO2201603.

45  Davis C, Naci H, Gurpinar E, *et al.* Availability of evidence of benefits on overall survival and quality of life of cancer drugs approved by European Medicines Agency: retrospective cohort study of drug approvals 2009-13. *BMJ*. 2017;359:4530.

# Figure 1: Immune related Adverse Events free survival

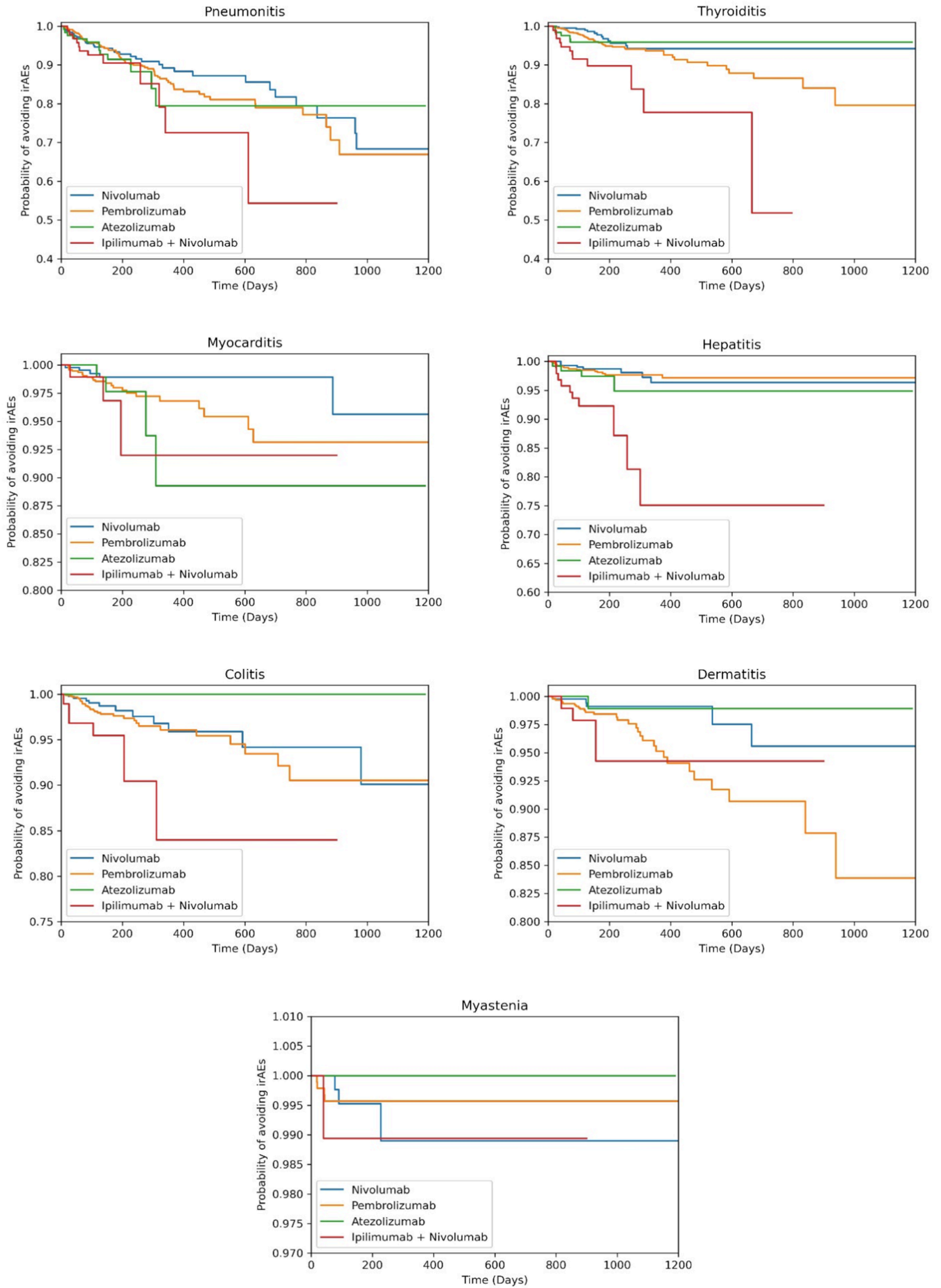

**Figure 2:** ROC-AUC results for the joint model for confirming a sentence contains positive reference of IrAEs attributed to ICIs

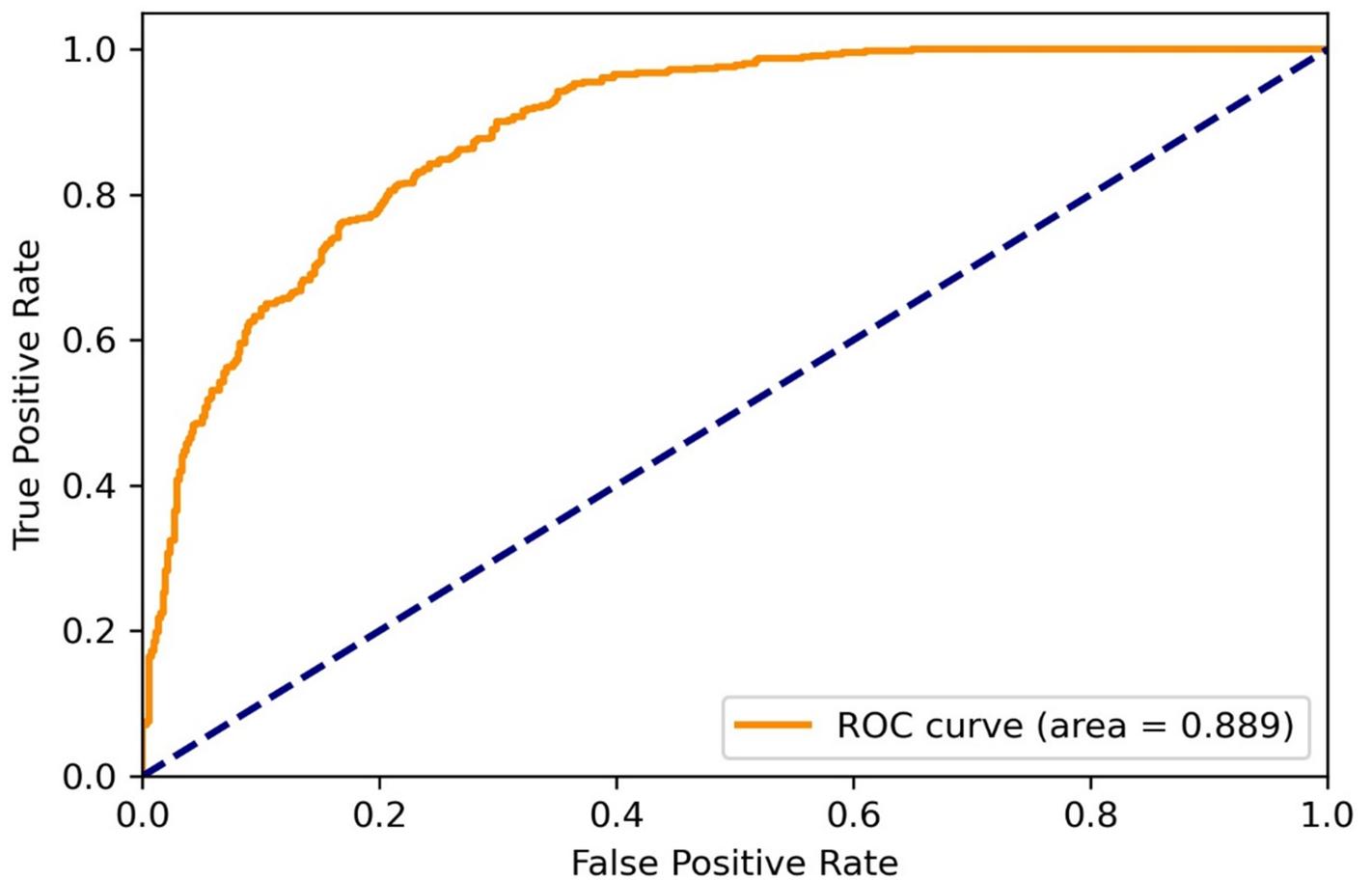

# Research pipeline diagram

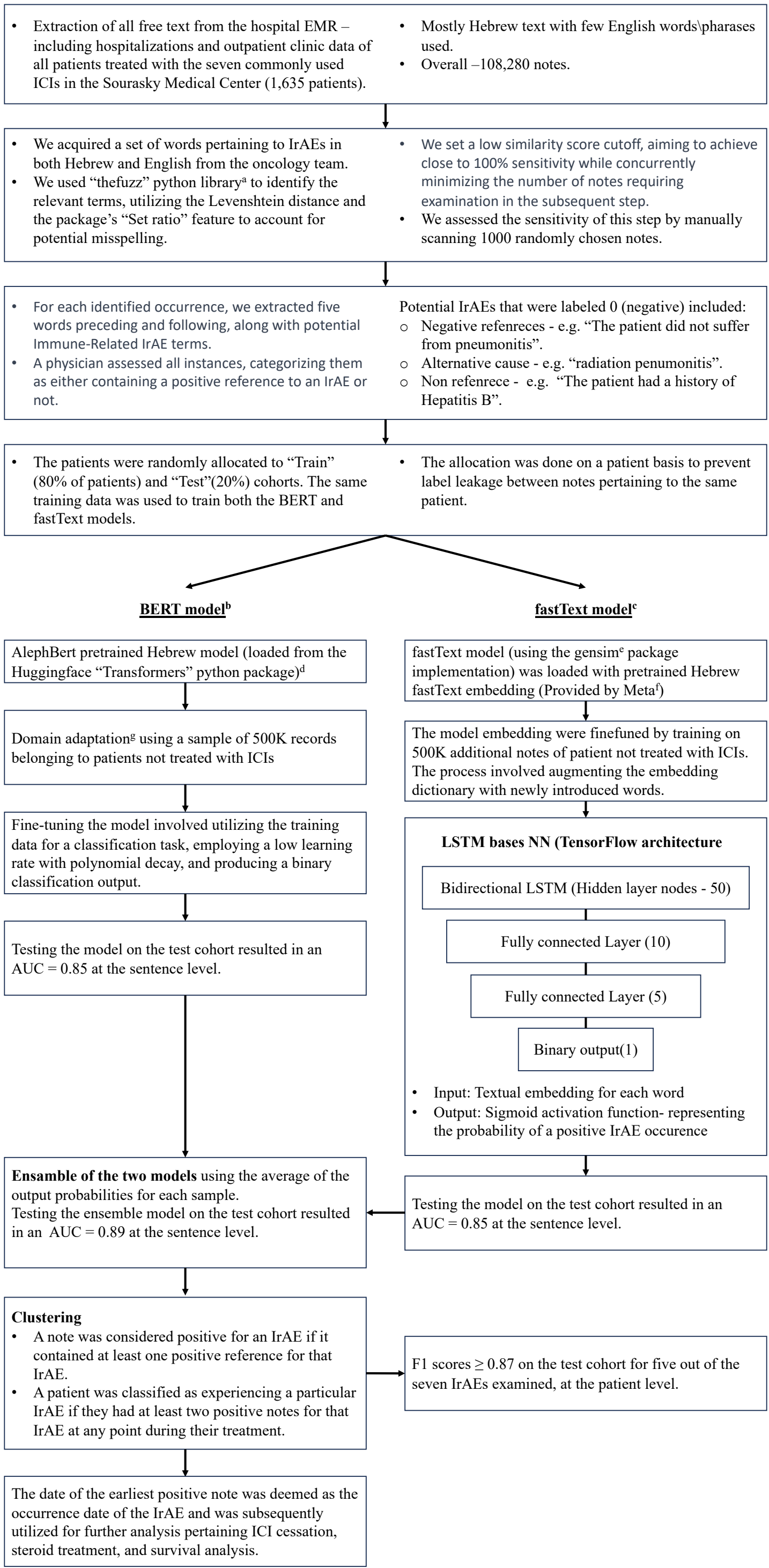

# Supplementary material

**Table S1:** IrAEs predicted by the study algorithm among the test cohort patients

|  | Actual | Predicted |
|---|---|---|
| Colitis | 9 | 16 |
| Dermatitis | 8 | 7 |
| Hepatitis | 8 | 10 |
| Myasthenia Gravis | 2 | 2 |
| Myocarditis | 8 | 8 |
| Pneumonitis | 33 | 36 |
| Thyroiditis | 12 | 11 |

**Table S2:** Discontinuation of ICI treatment following the first diagnosis of IrAEs

| Drug | N | Pneumonitis | Hepatitis | Thyroiditis | MG | Dermatitis | Colitis | Myocarditis |
|---|---|---|---|---|---|---|---|---|
| Atezolizumab | 122 | 4 (30.8%) | 2 (40%) | 0 | 0 | 0 | 0 | 1 (25%) |
| Avelumab | 11 | 0 | 0 | 0 | 0 | 0 | 0 | 0 |
| Durvalumab | 58 | 4 (33.3%) | 1 (100%) | 0 | 1 (100%) | 0 | 0 | 2 (66.7%) |
| Pembrolizumab | 926 | 23 (26.1%) | 7 (33.3%) | 6 (14%) | 3 (60%) | 5 (21.7%) | 8 (26.7%) | 7 (31.8%) |
| Nivolumab | 424 | 10 (27%) | 2 (28.6%) | 1 (9.1%) | 0 | 0 | 3 (33.3%) | 2 (40%) |
| Ipilimumab + Nivolumab | 94 | 1 (7.1%) | 2 (16.7%) | 2 (11.8%) | 0 | 1 (25%) | 0 | 1 (33.3%) |
| Mean steroid treatment rate (%) |  | 25.5% | 30.4% | 11.1% | 40.0% | 16.7% | 23.9% | 35.1% |

The percentage in brackets reflects the proportion of patients who stop ICI treatment among those who experienced IrAEs.

**Table S3:** Prevalence of new IrAEs during the additional follow-up period.

| Drug | Pneumonitis | Hepatitis | Thyroiditis | MG | Dermatitis | Colitis | Myocarditis |
|---|---|---|---|---|---|---|---|
| Atezolizumab | 2 | 0 | 1 | 0 | 1 | 1 | 0 |
| Avelumab | 1 | 1 | 0 | 0 | 0 | 0 | 0 |
| Durvalumab | 5 | 0 | 1 | 0 | 0 | 0 | 1 |
| Pembrolizumab | 17 | 2 | 9 | 2 | 5 | 13 | 6 |
| Nivolumab | 5 | 3 | 0 | 0 | 0 | 9 | 0 |
| Ipilimumab + Nivolumab | 5 | 1 | 3 | 1 | 1 | 1 | 0 |
| Mean steroid treatment rate (%) | 34.3% | 42.9% | 21.4% | 66.7% | 14.3% | 37.5% | 37.5% |